\documentclass[11pt]{article}

\usepackage[preprint]{acl}

\usepackage{times}
\usepackage{latexsym}
\usepackage{multirow}
\usepackage{amsmath}
\usepackage{amssymb}
\usepackage{amsfonts}
\usepackage[T1]{fontenc}

\usepackage[utf8]{inputenc}

\usepackage{microtype}

\usepackage{inconsolata}

\usepackage{graphicx}
\usepackage{booktabs}

\title{MOSLD-Bench: Multilingual Open-Set Learning and Discovery Benchmark for Text Categorization}

\author{
\textbf{Adriana-Valentina Costache}$^{*}$\textbf{, Daria-Nicoleta Dragomir}$^{*}$\textbf{, Silviu-Florin Gheorghe,}\\
\textbf{Eduard Poesina, Paul Irofti, Radu Tudor Ionescu}$^{\diamond}$\\
Department of Computer Science, University of Bucharest, Romania\\
$^{*}$Equal contribution. $^{\diamond}${\texttt{raducu.ionescu@gmail.com}}
}

\begin{document}
\maketitle
\begin{abstract}
Open-set learning and discovery (OSLD) is a challenging machine learning task in which samples from new (unknown) classes can appear at test time. It can be seen as a generalization of zero-shot learning, where the new classes are not known a priori, hence involving the active discovery of new classes. While zero-shot learning has been extensively studied in text classification, especially with the emergence of pre-trained language models, open-set learning and discovery is a comparatively new setup for the text domain. To this end, we introduce the first \textbf{m}ultilingual \textbf{o}pen-\textbf{s}et \textbf{l}earning and \textbf{d}iscovery (MOSLD) benchmark for text categorization by topic, comprising 960K data samples across 12 languages. To construct the benchmark, we (i) rearrange existing datasets and (ii) collect new data samples from the news domain. Moreover, we propose a novel framework for the OSLD task, which integrates multiple stages to continuously discover and learn new classes. We evaluate several language models, including our own, to obtain results that can be used as reference for future work. We release our benchmark at \url{https://github.com/Adriana19Valentina/MOSLD-Bench}.
\end{abstract}

\setlength{\abovedisplayskip}{5.5pt}
\setlength{\belowdisplayskip}{5.5pt}

\section{Introduction}

In machine learning, training and test data are presumed to be independently and identically sampled from the same distribution \cite{Vapnik-SB-1995}. However, this scenario is often violated in practice. Real-world data can exhibit one or multiple types of distribution shifts, rendering training and test data dissimilar. Various types of shifts lead to various task formalizations \cite{costache2025survey}, such as classification with background class \cite{Dhamija-NeurIPS-2018,Hendrycks-ICLR-2019,Lee-ICLR-2018,Liu-NeurIPS-2020}, zero-shot learning \cite{Gera-EMNLP-2019,Meng-EMNLP-2020,Meng-NIPS-2022,Sanh-ICLR-2022, Yin-EMNLP-2019, Zhang-ARXIV-2024}, and open-set learning \cite{Chen-ACL-2024, Geng-TPAMI-2020, Scheirer-TPAMI-2012}. 

Among these tasks, zero-shot learning (ZSL) is the most extensively studied, due to the spectacular advances obtained by pre-trained language models \cite{Bommasani-ARXIV-2021, Yang-CSUR-2025}. ZSL is achieved either by adapting various related tasks in order to solve the classification problem \cite{Yang-CSUR-2025,Yin-EMNLP-2019, Zhou-IJMLC-2024}, or by generating synthetic data and subsequently using a supervised method \cite{Hu-ICML-2018,Meng-NIPS-2022,Ye-EMNLP-2022,Yu-ACL-2023}. 
While ZSL received significant attention in recent literature, it requires class labels to be known a priori, i.e.~it does not cover the case where new classes that emerge during inference are completely unknown\footnote{Please see the distinction between Unknown Known Classes (UKCs) and Unknown
Unknown Classes (UUCs) made by \citet{Geng-TPAMI-2020}.}. 

\begin{table*}[t!]
\centering
\setlength\tabcolsep{0.3em}
\resizebox{\linewidth}{!}{%
\begin{tabular}{llcccccccccccc}
\toprule
\multirow{2}{*}{\textbf{Language}} & \multirow{2}{*}{\textbf{Source}} & \multicolumn{2}{c}{\textbf{Train}} & \multicolumn{2}{c}{\textbf{Val}} & \multicolumn{2}{c}{\textbf{Test 1}} & \multicolumn{2}{c}{\textbf{Test 2}} & \multicolumn{2}{c}{\textbf{Test 3}} & \multicolumn{2}{c}{\textbf{Total}} \\
 &  & {\#classes} & {\#samples} & {\#classes} & {\#samples} & {\#classes} & {\#samples} & {\#classes} & {\#samples} & {\#classes} & {\#samples} & {\#classes} & {\#samples} \\
\midrule
Arabic & Ultimate Arabic News & 4 & 98,350 & 4 & 14,341 & 6 & 18,341 & 8 & 24,953 & 10 & 32,664 & 10 & 188,649 \\
Bengali & L3Cube & 4 & 14,402 & 4 & 3,602 & 6 & 12,164 & 8 & 17,916 & 10 & 23,837 & 10 & 71,921 \\
Chinese & THUCNews & 4 & 22,367 & 4 & 7,998 & 7 & 13,555 & 10 & 25,174 & 14 & 52,346 & 14 & 121,440 \\
English & DBpedia & 4 & 40,000 & 4 & 4,000 & 7 & 21,000 & 10 & 30,000 & 14 & 42,000 & 14 & 137,000 \\
French & Collected & 4 & 42,534 & 4 & 5,962 & 6 & 11,255 & 8 & 17,998 & 10 & 21,388 & 10 & 99,137 \\
Hungarian & Collected & 4 & 10,323 & 4 & 1,141 & 6 & 7,527 & 8 & 12,250 & 10 & 16,943 & 10 & 48,184 \\
Italian & Collected & 4 & 10,238 & 4 & 1,809 & 6 & 5,936 & 8 & 7,753 & 10 & 9,730 & 10 & 35,466 \\
Japanese & Collected & 4 & 11,371 & 4 & 2,009 & 6 & 6,000 & 8 & 8,000 & 10 & 10,000 & 10 & 37,380 \\
Romanian & MOROCO+collected & 4 & 32,157 & 4 & 19,614 & 6 & 8,577 & 8 & 12,408 & 10 & 17,598 & 10 & 90,354 \\
Russian & Rus News & 4 & 11,248 & 4 & 2,813 & 6 & 8,204 & 8 & 11,673 & 10 & 15,781 & 10 & 49,719 \\
Spanish & Collected & 4 & 20,080 & 4 & 3,546 & 6 & 6,000 & 8 & 8,000 & 10 & 10,000 & 10 & 47,626 \\
Turkish & Kemik & 4 & 9,470 & 4 & 2,368 & 7 & 6,128 & 10 & 7,421 & 13 & 8,175 & 13 & 33,562 \\
\midrule
Total & Existing + collected & - & 322,540 & - & 69,203 & - & 124,687 & - & 183,546 & - & 260,462 & - & 960,438 \\
\bottomrule
\end{tabular}%
}
\vspace{-0.25cm}
\caption{Number of classes and number of samples for each language and each official subset in MOSLD-Bench.}
\label{tab:dataset_overview}
\vspace{-0.25cm}
\end{table*}

As highlighted by \citet{costache2025survey}, a more challenging task is \emph{open-set learning and discovery} (OSLD) \cite{Zheng-CVPR-2022}, where new classes, unseen during training, gradually appear in the test data. The goal is to detect these new classes and later learn to recognize them, without degrading the performance on the existing classes. OSLD can be seen as a relaxed case of \emph{open-set class incremental learning} \cite{Xu-ICCVW-2023}. \emph{Class incremental learning} (CIL) \cite{Kim-ANIPS-2022,Masana-TPAMI-2022,Zhou-TPAMI-2024} is a type of continual learning \cite{Liu-AI-2023}, where the problem is to learn a sequence of tasks, with each task adding new classes. 
Yet, a notable oversight in the realm of CIL is the inability to process the occurrence of unknown classes during inference. Open-set class incremental learning \cite{Xu-ICCVW-2023} merges class incremental learning and open-set recognition in a unified task formulation, where models must adapt to unknown classes, while access to already processed training data is restricted. In contrast, OSLD does not impose restricted access to training data, rendering catastrophic forgetting \cite{Kirkpatrick-PNAS-2017} easy to address. Nevertheless, OSLD is arguably one the most difficult versions of text classification under class distribution shift \cite{costache2025survey}, because the system must actively learn to identify new patterns in the data and adapt to them without any supervision. OSLD has practical motivation in text categorization, where new topics can emerge over time and systems need to handle them automatically. OSLD can also be seen as an extension of open-set learning (OSL), where samples from unknown classes have to be classified as such, in addition to being detected. While OSL has been explored in the text domain \cite{Chen-KDD-2023, Chen-ACL-2024, Kim-BigData-2022,Walkowiak-IntelliSys-2020,Walkowiak-IEA-2019, Walkowiak-ICAISC-2019}, the more challenging OSLD problem remains largely unexplored in NLP literature \cite{costache2025survey}.

To this end, we introduce the first \textbf{m}ultilingual \textbf{o}pen-\textbf{s}et \textbf{l}earning and \textbf{d}iscovery (MOSLD) benchmark for text categorization by topic. 
We first construct a comprehensive multilingual dataset via (i) rearranging existing datasets (such that some classes are not represented in the training set) and (ii) collecting new data samples from the news domain (for languages where existing resources are scarce). 
We further benchmark several small (e.g.~BERT \cite{Devlin-NAACL-2019}) and large (e.g.~GPT-4o \cite{Hurst-Arxiv-2024}) language models on MOSLD-Bench. Since the task was not previously explored, we propose a novel OSLD framework that integrates multiple stages (e.g.~outlier detection, k-means clustering, TFIDF-based keyword extraction, BERT retraining with pseudo-labeling) to continuously discover and learn new classes.

In summary, our contribution is twofold:
\begin{itemize}
    \item \vspace{-0.15cm} We introduce a novel multilingual dataset for open-set learning and discovery, which is obtained via new data gathering and existing dataset restructuring.
    \item \vspace{-0.2cm} We propose a new framework designed to address all the challenges of OSLD via a multi-stage processing pipeline based on keyword extraction, clustering, pseudo-labeling and model retraining.
\end{itemize}

\vspace{-0.2cm}
\section{Dataset} \label{sec:dataset}
\vspace{-0.1cm}

\noindent\textbf{Task definition.} 
Let $\mathcal{D}\!=\!\{(x, y) \mid x\!\in\!\mathcal{X}, y\!\in\!\mathcal{Y}\}$ be a training set, where $x$ is data sample from the space $\mathcal{X}$, and $y$ is a label from the space $\mathcal{Y}$. The goal of OSLD is to predict the labels for some test set $\mathcal{T}_i\!=\!\{(x, y) \mid x\!\in\!\mathcal{X}, y\!\in\!\mathcal{Y}_i^+\}$, where $i$ represents the test set index, which increases over time, and $\mathcal{Y}_i^+$ contains class labels that are not in $\mathcal{Y}$, i.e. $\mathcal{Y} \subset \mathcal{Y}_i^+$. At any timestep $i$, the unlabeled test sets $\mathcal{T}_1, ..., \mathcal{T}_{i}$ and the original training set $\mathcal{D}$ can be used to obtain the model $h_i: \mathcal{X} \to \mathcal{Y}_i^+$, which is supposed to correctly identify samples belonging to $\mathcal{Y}$, to discover classes from $\mathcal{Y}_i^+ \setminus \mathcal{Y}$, and recognize samples from the discovered classes.

\noindent\textbf{Overview.} Following the task definition above, we construct a multilingual dataset for OSLD, such that for each language, only a subset of the total number of classes is included in the training data, while the remaining classes are gradually inserted during testing. More precisely, for each language, we construct three test sets, such that $\mathcal{Y}_i^+ \subset \mathcal{Y}_{i+1}^+, \forall i \in \{1, 2\}$. The dataset contains 960,438 samples from 12 typologically diverse languages, covering multiple language families and diverse writing systems (see Table \ref{tab:dataset_overview}). 

\noindent\textbf{Data collection.} The corpus is compiled from two complementary sources. For high-resource languages, we integrate publicly available datasets:  Ultimate Arabic News Dataset \cite{Al-Dulaimi-Mendeley-2022} for Arabic, L3Cube-IndicNews \cite{Mirashi-ICON-2023} for Bengali, DBpedia Ontology Dataset \cite{Zhang-NIPS-2016} for English, THUCNews \cite{Sun-THUNLP-2016} for Chinese, Turkish News Kemik \cite{Yildirim-PUJEES-2018} for Turkish, and Rus News Dataset \cite{Kuznetsov-HF-2024} for Russian. For Romanian, we extend the MOROCO dataset \cite{Butnaru-ACL-2019} (originally containing 6 categories) by collecting additional news articles for 4 new categories, resulting in a total of 10 categories. For languages lacking suitable public resources (French, Italian, Japanese, Hungarian, and Spanish), we collect news articles directly from local news portals. The collected data is subsequently anonymized to remove personally identifiable information. Category labels were derived from the existing editorial categories used by each news website, requiring no manual annotation. In the end, each language subset contains between 10 and 14 categories covering common news domains, such as politics, sport, economy, technology, culture, health, etc.

\noindent\textbf{Dataset organization.} To enable standardized evaluation, we apply a uniform splitting protocol across all languages. For each language, we designate 4 classes as ``known'' from the beginning (belonging to $\mathcal{Y}$), splitting their samples into training, validation, and test. The remaining classes are treated as ``unknown'' (belonging to $\mathcal{Y}_i^+ \setminus \mathcal{Y}$). These unknown classes are progressively introduced across three evaluation stages. Test $\mathcal{T}_1$ contains test samples from the 4 baseline classes plus samples from 2-3 unknown classes. Test $\mathcal{T}_2$ contains test samples from the baseline classes, samples from the classes introduced in $\mathcal{T}_1$, plus samples from another 2-3 new classes. Test $\mathcal{T}_3$ follows the same pattern, incorporating the rest of unknown classes. Once a class is introduced, it remains present in all subsequent test sets. This structure enables evaluation of both the model's ability to maintain accuracy on known classes and its capacity to detect novel, previously unseen categories. In Table~\ref{tab:dataset_overview}, we provide detailed information about the number of classes and the number of samples per language and per subset, respectively.

\noindent\textbf{Evaluation procedure.} We evaluate models on two complementary objectives: maintaining performance on previously learned classes, and performing well on novel categories. We report accuracy rates and macro-averaged F1 scores for three samples groups: (1) \textbf{overall} -- includes all test samples in $\mathcal{T}_i$, (2) \textbf{known} -- includes samples from classes seen during training, (3) \textbf{unknown} -- includes samples from classes introduced in the current test set $\mathcal{T}_i$, for all $i \in \{1,2,3\}$. We track accuracy on each group across successive stages, measuring how performance on earlier classes degrades as new categories are learned. Since discovered classes are learned in an unsupervised manner, without access to ground-truth labels, we determine the correspondence between discovered and ground-truth labels using Hungarian matching. More details about the matching procedure are given in Appendix \ref{sec_matching_classes}.

\begin{table*}[ht]
\centering
\setlength\tabcolsep{0.35em}
\resizebox{\linewidth}{!}{%
\begin{tabular}{lccc|ccc|ccc}
\toprule
& \multicolumn{3}{c|}{\textbf{Test 1}} & \multicolumn{3}{c|}{\textbf{Test 2}} & \multicolumn{3}{c}{\textbf{Test 3}} \\
\textbf{Language} & \textbf{Baseline} & \textbf{V1} & \textbf{V2} & \textbf{Baseline} & \textbf{V1} & \textbf{V2} & \textbf{Baseline} & \textbf{V1} & \textbf{V2} \\
& Acc / F1 & Acc / F1 & Acc / F1 & Acc / F1 & Acc / F1 & Acc / F1 & Acc / F1 & Acc / F1 & Acc / F1 \\
\midrule
Arabic & 0.517 / 0.371 & \textbf{0.906} / \textbf{0.900} & 0.902 / 0.896 & 0.370 / 0.226 & 0.679 / 0.706 & \textbf{0.680} / \textbf{0.707} & 0.285 / 0.154 & 0.533 / 0.508 & \textbf{0.541} / \textbf{0.518} \\
Bengali & 0.448 / 0.301 & \textbf{0.579} / \textbf {0.595} & 0.577 / 0.592 & 0.302 / 0.160 & \textbf{0.453} / \textbf{0.469} & 0.448 / 0.461 & 0.226 / 0.105 & 0.334 / 0.351 & \textbf{0.338} / \textbf{0.351} \\
Chinese & 0.396 / 0.233 & 0.725 / 0.742 & \textbf{0.729} / \textbf{0.746} & 0.207 / 0.082 & 0.635 / 0.634 & \textbf{0.648} / \textbf{0.639} & 0.099 / 0.019 & \textbf{0.418} / \textbf{0.422} & 0.408 / 0.422 \\
English & 0.570 / 0.467 & \textbf{0.727} / \textbf{0.695} & 0.722 / {0.691} & 0.399 / 0.301 & 0.515 / 0.481 & \textbf{0.578} / \textbf{0.532} & 0.285 / 0.165 & 0.430 / 0.363 & \textbf{0.432} / \textbf{0.369} \\
French & 0.340 / 0.178 & \textbf{0.889} / \textbf{0.874}& 0.862 / 0.845 & 0.213 / 0.079 & 0.577 / 0.626 & \textbf{0.586} / \textbf{0.630} & 0.179 / 0.057 & {0.410} / {0.448} & \textbf{0.427} / \textbf{0.458} \\
Hungarian & 0.512 / 0.364 & 0.682 / 0.715 & \textbf{0.720} / \textbf{0.742} & 0.314 / 0.160 & \textbf{0.647} / \textbf{0.666} & 0.543 / 0.577 & 0.227 / 0.093 & 0.439 / 0.512 & \textbf{0.544} / \textbf{0.562} \\
Italian & 0.632 / 0.529 & 0.733 /  0.710 & \textbf{0.737} / \textbf{0.717} & 0.482 / 0.351 & 0.578 / 0.539 & \textbf{0.579} / \textbf{0.542} & 0.386 / 0.243 & \textbf{0.473} / 0.436 & 0.469 / \textbf{0.437} \\
Japanese & 0.622 / 0.508 & 0.626 / \textbf{0.606} & \textbf{0.627} / 0.602 &  0.470 / 0.332 & 0.623 / 0.592 & \textbf{0.639} / \textbf{0.603} & 0.374 / 0.232 & 0.507 / 0.498 & \textbf{0.529} / \textbf{0.510} \\
Romanian & \textbf{0.452} / 0.370 & 0.443 / 0.526 & \textbf{0.452}/ \textbf{0.529}& 0.312 / 0.203 & 0.332 / 0.402 & \textbf{0.337} / \textbf{0.403} & 0.219 / 0.137 & \textbf{0.220} / \textbf{0.291} & 0.222 / 0.295 \\
Russian & 0.526 / 0.383 & \textbf{0.658} / \textbf{0.679} & 0.649 / 0.670 & 0.366 / 0.210 & \textbf{0.555} / \textbf{0.544} & 0.522 / 0.534 & 0.274 / 0.128 & \textbf{0.407} / \textbf{0.419} & 0.400 / 0.411 \\
Spanish & \textbf{0.640} / 0.522 & 0.639 / \textbf{0.608} & 0.630 / 0.598 & 0.480 / 0.335 & \textbf{0.542} / \textbf{0.490} & 0.537 / 0.488 & 0.385 / 0.239 & 0.436 / 0.368 & \textbf{0.444} / \textbf{0.379} \\
Turkish & \textbf{0.518} / \textbf{0.434} & 0.507 / 0.376 & 0.512 / 0.379 & \textbf{0.429} / \textbf{0.328} & 0.414 / 0.265 & 0.423 / 0.270 & \textbf{0.393} / \textbf{0.289} & 0.362 / 0.194 & 0.361 / 0.196 \\
\bottomrule
\end{tabular}
}
\vspace{-0.25cm}
\caption{\textbf{Overall} accuracy and F1 scores across 12 languages for each evaluation stage in MOSLD-Bench. The best score for each language and each test set is highlighted in bold.}
\vspace{-0.25cm}
\label{tab:results_overall}
\end{table*}

\vspace{-0.1cm}
\section{Methods} \label{sec:methods}
\vspace{-0.1cm}

\noindent\textbf{Baseline based on known-class supervision.}
To establish a performance lower bound, we fine-tune language-specific pre-trained BERT encoders. The specific version of BERT for each language is reported in Table \ref{tab:berts} from Appendix \ref{sec_berts}. The models are optimized via standard supervised learning (cross-entropy loss) only on the initial set of known classes $\mathcal{Y}$. The weights stay frozen after the initial training phase, while the models are evaluated sequentially on all three test sets. This setting corresponds to a classifier that does not adapt to distribution and label-space shift. The corresponding performance levels serve as a reference point for OSLD methods that learn continuously.

\noindent\textbf{Proposed OSLD methods.} We propose two alternative OSLD methods that follow a joint processing pipeline comprising multiple stages: (1) outlier sample detection, (2) outlier data clustering, (3) class-specific keyword extraction, (4) model retraining. These steps are sequentially applied on each test set $\mathcal{T}_i$.
We start from the language-specific BERT models fine-tuned on the training set (comprising only known classes). In step (1), we identify samples not belonging to any learned (known) category using energy-based detection. The energy score of input $x$ is computed as:
\begin{equation}
E(x) = -\log \left( \sum_{j=1}^{N}\exp\left(h^j(x)\right) \right),
\end{equation}
where $h^j(x)$ denotes the $j$-th logit output of the classifier $h$, and $N$ is the number of currently known classes. Intuitively, samples from known classes yield lower energy scores due to higher confidence predictions, while outliers produce higher energy values. Therefore, samples with a high energy (top $15\%$) are labeled as outliers, while the remaining samples are assigned to one of the known classes. In step (2), we cluster outliers using k-means on \texttt{[CLS]} embeddings provided by the language model. The optimal $k$ is determined by maximizing the silhouette coefficient. The resulting $k$ represents the number of discovered classes. In step (3), we extract a list of representative keywords for each cluster, which can be further used to match the respective cluster with one of the ground-truth classes. To extract keywords, we merge all the samples in each cluster into a single document, then apply the TFIDF scheme over the resulting documents. This promotes words that appear in only one of the clusters. We select the top 10 keywords for each cluster. Each keyword list is passed through BERT and the corresponding \texttt{[CLS]} embedding becomes a centroid for the corresponding cluster. In step (4), we sort the samples in each cluster based on the cosine similarity with their centroid. We keep $40\%$ of samples closest to each centroid for model retraining. At this stage, we implement two alternative approaches. In the first approach (V1), the set of classes is expanded with the discovered clusters and the model is retrained with standard cross-entropy. In the second approach (V2), we augment cross-entropy with a contrastive term that pulls sample embeddings towards their centroid, grounding representations in semantic information from clustering:
\begin{equation}\label{eq_total_loss}
\mathcal{L} = \mathcal{L}_{\text{CE}} + \lambda \cdot \mathcal{L}_{\text{CL}}.
\end{equation}
In Eq.~\eqref{eq_total_loss}, $\mathcal{L}_{\text{CE}}$ is the cross-entropy and $\mathcal{L}_{\text{CL}}$ is defined as:
\begin{equation}\label{eq_contrast_loss}
\mathcal{L}_{\text{CL}}\!=\! -\frac{1}{|\mathcal{U}|}\!\sum_{i \in \mathcal{U}} \log\!\left(\!\frac{\exp(\operatorname{sim}(\mathbf{e}_i, \mathbf{c}_{\hat{y}_i}) / \tau)}{\sum_{j=1}^{k} \exp(\operatorname{sim}(\mathbf{e}_i, \mathbf{c}_j) / \tau)} \!\right)\!\!,
\end{equation}
where $\mathcal{U}$ denotes the set of samples from discovered classes, $\mathbf{e}_i$ is the \texttt{[CLS]} embedding of sample $x_i$, $\mathbf{c}_{\hat{y}_i}$ is the centroid of the cluster ${\hat{y}_i}$ that includes sample $x_i$, $\operatorname{sim}(\cdot, \cdot)$ is the cosine similarity, and $\tau$ is a temperature parameter.

\vspace{-0.2cm}
\section{Experiments}
\vspace{-0.1cm}

\noindent\textbf{Hyperparameter tuning.} To reproduce results, we provide details about hyperparameter choices in Appendix \ref{sec_hyper}.

\noindent\textbf{Baseline vs.~OSLD BERT models.} In Table \ref{tab:results_overall}, we present the accuracy and F1 scores for all classes (\textbf{overall}), across all languages and evaluation stages. The baseline model, which remains frozen after initial training on known classes, shows consistent performance degradation as new classes are introduced. This is expected, since the baseline cannot recognize samples from unknown categories, assigning them to one of the known classes. In contrast, the proposed OSLD methods (V1 and V2) demonstrate the ability to discover and learn new classes, generally achieving substantially higher scores than the baseline. Comparing V1 and V2, we observe that integrating the contrastive loss (V2) provides improvements on several languages, particularly in early stages (T1 and T2). We present results for \textbf{known} and \textbf{unknown} classes in Tables \ref{tab:results_known} and \ref{tab:results_unknown}, discussing them in Appendix \ref{sec_known_vs_unknown}.

\noindent\textbf{Small vs.~large language models.} We discuss this comparative study in Appendix \ref{sec_small_vs_large}.

\vspace{-0.2cm}
\section{Conclusion}
\vspace{-0.1cm}
We introduced a multilingual dataset to benchmark methods for open-set learning and discovery. To construct the dataset, we adopted two approaches: (i) reorganizing existing datasets, and (ii) collecting new data from the web. We evaluated a set of baseline methods, creating reference points for future research on the topic. Since OSLD is a new task in the NLP domain, we also proposed a method specifically designed to solve the OSLD task. 

In future work, we aim to evaluate additional OSLD methods on MOSLD-Bench.

\section*{Acknowledgments}
This work was supported by a grant of the Ministry of Research, Innovation and Digitization, CCCDI - UEFISCDI, project number PN-IV-P6-6.3-SOL-2024-0090, within PNCDI IV.

\section*{Limitations}

Our evaluation includes a limited number of baseline models, reflecting the scarcity of methods explicitly designed for open-set learning and discovery. According to \citet{costache2025survey}, OSLD has not been explored in the NLP domain. We aim to address this limitation by closely tracking future work on OSLD and include newly developed models.

In addition, the chosen large language model, GPT-4o, is evaluated in a reduced setup, comprising two languages and only one test stage. This is due to the high cost of the API-based inference with GPT-4o.

\bibliography{custom}

\appendix

\begin{table}[t!]
\centering
\setlength\tabcolsep{0.3em}
\begin{tabular}{ll}
\toprule
{\textbf{Language}} & {\textbf{Model}} \\
\midrule
Arabic & AraBERT \cite{antoun2020arabert}  \\
Bengali & BanglaBERT \cite{Sagor_2020} \\
Chinese & Chinese BERT \cite{Devlin-NAACL-2019} \\
English & BERT \cite{Devlin-NAACL-2019} \\
French & French BERT \cite{Schweter-EuropeanaBERT-2020} \\
Hungarian & huBERT \cite{Nemeskey-huBERT-2021}\\
Italian & Italian BERT \cite{Schweter-EuropeanaBERT-2020}\\
Japanese & TohokuBERT \cite{tohoku2023bertbasejapanesev3} \\
Romanian & RoBERT\cite{masala2020robert} \\
Russian & ruBERT \cite{zmitrovich2023family} \\
Spanish & Spanish BERT \cite{CaneteCFP2020} \\
Turkish & BERTurk \cite{schweter2020berturk} \\
\bottomrule
\end{tabular}%
\vspace{-0.2cm}
\caption{All models use the base architecture, adapted for uncased text processing.}
\label{tab:berts}
\vspace{-0.2cm}
\end{table}

\section{Appendix}
\label{sec:appendix}

\vspace{-0.1cm}
\subsection{Discovered and Ground-Truth Class Matching}
\label{sec_matching_classes}
\vspace{-0.1cm}

\begin{table*}[ht]
\centering
\resizebox{\linewidth}{!}{%
\begin{tabular}{lccc|ccc|ccc}
\toprule
 & \multicolumn{3}{c|}{\textbf{Test 1}} & \multicolumn{3}{c|}{\textbf{Test 2}} & \multicolumn{3}{c}{\textbf{Test 3}} \\
\textbf{Language} & \textbf{Baseline} & \textbf{V1} & \textbf{V2} & \textbf{Baseline} & \textbf{V1} & \textbf{V2} & \textbf{Baseline} & \textbf{V1} & \textbf{V2} \\
 & Acc / F1 & Acc / F1 & Acc / F1 & Acc / F1 & Acc / F1 & Acc / F1 & Acc / F1 & Acc / F1 & Acc / F1 \\
\midrule
Arabic &  \textbf{0.988} / \textbf{0.988}  & 0.959 / 0.959 & 0.957 / 0.958 &  \textbf{0.986} / \textbf{0.986} & 0.929 / 0.941 & 0.929 / 0.942 & \textbf{0.985} / \textbf{0.985}  & 0.952 / 0.957 & 0.950 / 0.955 \\
Bengali & \textbf{0.908} / \textbf{0.908} & 0.805 / 0.816 & 0.815 / 0.832 & \textbf{0.902} / \textbf{0.901} & 0.796 / 0.818 & 0.800 / 0.828 & \textbf{0.900} / \textbf{0.900} & 0.797 / 0.825 & 0.777 / 0.814 \\
Chinese & \textbf{0.989} / \textbf{0.989} & 0.957 / 0.973 & 0.968 / 0.978 & \textbf{0.989} / \textbf{0.989} & 0.965 / 0.978 & 0.969 / 0.980 & \textbf{0.989} / \textbf{0.989} & 0.944 / 0.967 & 0.945 / 0.966 \\
English & \textbf{0.998} / \textbf{0.998} & 0.931 / 0.960 & 0.931 / 0.960 & \textbf{0.998} / \textbf{0.998} & 0.907 / 0.943 & 0.917 / 0.950 & \textbf{0.999} / \textbf{0.999} & 0.909 / 0.944 & 0.908 / 0.944 \\
French & \textbf{0.963} / \textbf{0.963} & 0.945 / 0.951 & 0.957 / 0.957 & \textbf{0.966} / \textbf{0.966} & 0.932 / 0.949 & 0.936 / 0.951 & \textbf{0.967} / \textbf{0.967} & 0.921 / 0.945 & 0.935 / 0.949 \\
Hungarian & \textbf{0.972} / \textbf{0.972} & 0.947 / 0.957 & 0.951 / 0.995 & \textbf{0.971} / \textbf{0.971} & 0.932 / 0.950 & 0.944 / 0.950 & \textbf{0.972} / \textbf{0.972} & 0.931 / 0.950 & 0.925 / 0.947 \\
Italian & \textbf{0.938} / \textbf{0.938} & 0.882 / 0.910 & 0.896 / 0.913 & \textbf{0.935} / \textbf{0.935} & 0.841 / 0.890 & 0.842 / 0.888 & \textbf{0.939} / \textbf{0.938} & 0.812 / 0.877 & 0.832 / 0.886 \\
Japanese & \textbf{0.933} / \textbf{0.933} & 0.902 / 0.926 & 0.904 / 0.927 & \textbf{0.941} / \textbf{0.941} & 0.841 / 0.899 & 0.872 / 0.917 & \textbf{0.936} / \textbf{0.936} & 0.781 / 0.862 & 0.825 / 0.890 \\
Romanian & \textbf{0.970} / \textbf{0.970} & 0.948 / 0.962 & 0.970 / 0.970 & \textbf{0.969} / \textbf{0.969} & 0.947 / 0.962 & 0.959 / 0.963 & \textbf{0.967} / \textbf{0.967} & 0.944 / 0.959 & 0.944 / 0.959 \\
Russian & \textbf{0.921} / \textbf{0.921} & 0.888 / 0.911 & 0.890 / 0.907 & \textbf{0.913} / \textbf{0.912} & 0.841 / 0.889 & 0.845 / 0.891 & \textbf{0.922} / \textbf{0.922} & 0.820 / 0.880 & 0.818 / 0.877 \\
Spanish & \textbf{0.961} / \textbf{0.961} & 0.932 / 0.948 & 0.925 / 0.945 & \textbf{0.960} / \textbf{0.960} & 0.928 / 0.947 & 0.922 / 0.942 & \textbf{0.962} / \textbf{0.962} & 0.928 / 0.949 & 0.913 / 0.942 \\
Turkish & \textbf{0.805} / \textbf{0.799} & 0.778 / 0.775 & 0.789 / 0.777 & \textbf{0.806} / \textbf{0.801} & 0.741 / 0.752 & 0.762 / 0.763 & \textbf{0.814} / \textbf{0.809} & 0.722 / 0.746 & 0.718 / 0.743 \\
\bottomrule
\end{tabular}
 }
\vspace{-0.2cm}
\caption{Accuracy and F1 scores computed of \textbf{known} classes across 12 languages for each evaluation stage in MOSLD-Bench. The best score for each language and each test set is highlighted in bold.}
\label{tab:results_known}
\end{table*}

\begin{table*}[ht]
\centering
\resizebox{\linewidth}{!}{%
\begin{tabular}{lccc|ccc|ccc}
\toprule
 & \multicolumn{3}{c|}{\textbf{Test 1}} & \multicolumn{3}{c|}{\textbf{Test 2}} & \multicolumn{3}{c}{\textbf{Test 3}} \\
\textbf{Language} & \textbf{Baseline} & \textbf{V1} & \textbf{V2} & \textbf{Baseline} & \textbf{V1} & \textbf{V2} & \textbf{Baseline} & \textbf{V1} & \textbf{V2} \\
 & Acc / F1 & Acc / F1 & Acc / F1 & Acc / F1 & Acc / F1 & Acc / F1 & Acc / F1 & Acc / F1 & Acc / F1 \\
\midrule
Arabic & 0 / 0 & \textbf{0.845} / \textbf{0.904} & {0.838} / {0.901} & 0 / 0 & {0.523} / {0.532} & \textbf{0.525} / \textbf{0.534} & 0 / 0 & 0.254 / 0.260 & \textbf{0.266} / \textbf{0.272} \\

Bengali & 0 / 0 & \textbf{0.346} /  \textbf{0.449} & 0.205 / 0.086 & 0 / 0 & \textbf{0.281} / \textbf{0.346} & 0.270 / 0.326 & 0 / 0 & 0.178 / 0.231 & \textbf{0.190} / \textbf{0.240} \\

Chinese & 0 / 0 & \textbf{0.572} / 0.481 & 0.569 / \textbf{0.482} & 0 / 0 & 0.548 / {0.446} & \textbf{0.562} / \textbf{0.462} & 0 / 0 & \textbf{0.359} / {0.317} & 0.348 / \textbf{0.319} \\

English & 0 / 0 & \textbf{0.455} / \textbf{0.379} & 0.444 / {0.369} & 0 / 0 & 0.253 / 0.185 & \textbf{0.351} /\textbf{ 0.268} & 0 / 0 & 0.239 / 0.142 & \textbf{0.242} / \textbf{0.153} \\

French & 0 / 0 & \textbf{0.859} / \textbf{0.918} & 0.810 / 0.892 & 0 / 0 & 0.476 / 0.511 & \textbf{0.487} / \textbf{0.523} & 0 / 0 & 0.293 / 0.282 & \textbf{0.311} / \textbf{0.302} \\

Hungarian & 0 / 0 & 0.388 / 0.344 & \textbf{0.462} /\textbf{ 0.446} & 0 / 0 & \textbf{0.510} / \textbf{0.441} & 0.351 / 0.270 & 0 / 0 & 0.228 / 0.288 & \textbf{0.427} / \textbf{0.359} \\

Italian & 0 / 0 & \textbf{0.426} / {0.450} & 0.408 / \textbf{0.482} & 0 / 0 & 0.298 / 0.269 & \textbf{0.299} / \textbf{0.276} & 0 / 0 & \textbf{0.236} / 0.216 & 0.215 / \textbf{0.217} \\

Japanese & 0 / 0 & \textbf{0.075} / \textbf{0.090} & 0.074 / \textbf{0.090} & 0 / 0 & \textbf{0.406} / 0.363 & 0.405 / \textbf{0.367} & 0 / 0 & 0.325 / 0.301 & \textbf{0.332} / \textbf{0.306} \\

Romanian & 0 / 0 & \textbf{0.002} / \textbf{0.004} & \textbf{0.002} / 0.001 & 0 / 0 & 0.040 / 0.089 & \textbf{0.042} / \textbf{0.093}& {0} / {0} & {0.007} / {0.019} & \textbf{0.010} / \textbf{0.025} \\

Russian & 0 / 0 & \textbf{0.353} / \textbf{0.297} & 0.328 / 0.284 & 0 / 0 & \textbf{0.363} / \textbf{0.273} & 0.305 / {0.246} & 0 / 0 & \textbf{0.232} / \textbf{0.214} & 0.224 / 0.201 \\

Spanish & 0 / 0 & \textbf{0.053} / \textbf{0.061} & 0.040 / 0.048 & 0 / 0 & \textbf{0.156} / \textbf{0.142} & {0.153} / \textbf{0.142} & 0 / 0 & 0.108 / 0.076 & \textbf{0.131} / \textbf{0.092} \\

Turkish & 0 / 0 & \textbf{0.102} / 0.017 & 0.013 / \textbf{0.022} & 0 / 0 & \textbf{0.043} / 0.060 & 0.039 / \textbf{0.064} & 0 / 0 & 0.025 / 0.032 & \textbf{0.027} / \textbf{0.035} \\
\bottomrule
\end{tabular}
}
\vspace{-0.2cm}
\caption{Accuracy and F1 scores computed of \textbf{unknown} classes across 12 languages for each evaluation stage in MOSLD-Bench. The best score for each language and each test set is highlighted in bold.}
\label{tab:results_unknown}
\vspace{-0.2cm}
\end{table*}

To determine the correspondence between discovered and ground-truth labels, we compute the semantic similarity between discovered and ground-truth class names. For each discovered class, we compute an embedding representation from its characteristic keywords (a list of representative words returned by the OSLD method). We then measure cosine similarity between embeddings of discovered and ground-truth class names. The embeddings are obtained from language-specific pre-trained transformer models (see Table \ref{tab:berts}). The matching is performed exclusively during the evaluation stage to assess discovery quality, while ground-truth labels remain unused during training or inference. The optimal one-to-one assignment between discovered and ground-truth classes is obtained using Hungarian matching, enabling a fully automated evaluation, without any manual intervention.

\vspace{-0.1cm}
\subsection{Language-Specific BERT Models}
\label{sec_berts}
\vspace{-0.1cm}

In Table \ref{tab:berts}, we specify the BERT version used as the starting point for the fine-tuning process, for each language. Preliminary experiments with multilingual BERT (mBERT) \cite{Devlin-NAACL-2019} led to significantly worse results in several languages. To address this problem, we selected language-specific BERT models. The same models are used for both the naive approach (which ignores new classes) and the proposed OSLD approaches, ensuring a fair comparison between methods.

\vspace{-0.1cm}
\subsection{Hyperparameter Tuning}
\label{sec_hyper}
\vspace{-0.1cm}

We use consistent hyperparameters across all languages to ensure a fair comparison. All models are fine-tuned using the AdamW optimizer \cite{loshchilov2019decoupled} with a learning rate of $2 \cdot 10^{-5}$, weight decay of 0.01, and linear warmup over the first 100 steps. We train for 5 epochs with a batch size of 16.  During clustering, we search for the optimal $k$ between $2$ and $8$. We set $\lambda=0.3$ in Eq.~\eqref{eq_total_loss}, and $\tau=0.07$ in Eq.~\eqref{eq_contrast_loss}. We use the default values for all other hyperparameters.

\vspace{-0.1cm}
\subsection{Performance on Known vs.~Unknown Classes}
\label{sec_known_vs_unknown}
\vspace{-0.1cm}

In Tables \ref{tab:results_known} and \ref{tab:results_unknown}, we provide a detailed analysis of model performance, separating \textbf{known} classes (seen during training) from \textbf{unknown} classes (discovered during testing). 

\noindent
\textbf{Known classes.} 
The baseline model achieves high accuracy on known classes across all languages and evaluation stages, typically exceeding $90\%$. This confirms that the frozen baseline effectively retains its initial knowledge, which is expected when no model updates occur. Both V1 and V2 maintain competitive performance on known classes on the first test stage ($\mathcal{T}_1$), with only minor drops  with respect to the baseline. For subsequent performance stages, performance gaps become more evident for certain languages.

\begin{table*}[ht]
\centering
\small
\begin{tabular}{p{0.93\linewidth}}
\toprule
\textbf{(a) Prompt for label generation} \\
\midrule
You must behave as an Open-Set Classification Model which discovers the class of the presented texts. Each text has only one category. You will be provided the texts, numbered by their id. The known text classes are: \texttt{<LIST\_OF\_CLASS\_NAMES>}. \\[0.5em]
Your task is to discover how many new classes you can identify and name them alongside the existing ones. You must respond only with an array of all the existing classes. Do not use third party tools or python code to identify the names of the classes. Use only predictions based on what you read. Output only the classes. \\
\midrule
\textbf{(b) Prompt for sample classification} \\
\midrule
You need to behave as a text classification model. \\[0.5em]
You need to use the following classes in order to classify the texts: \texttt{<LIST\_OF\_CLASS\_NAMES>} \\[0.5em]
For each text example, output ONLY the class id (a single number from 0--\texttt{<NUM\_OF\_CLASSES>}). Nothing else. \\
\bottomrule
\end{tabular}
\vspace{-0.2cm}
\caption{Prompts used for GPT-4o: (a) open-set discovery and generation of class labels, and (b) classification of text samples. Text files containing training and test samples were attached to both prompts.}
\label{tab:prompts_gpt4o}
\end{table*}

\noindent
\textbf{Unknown classes.} The baseline trivially scores zero on unknown classes, as it lacks the capacity to predict categories outside its training distribution. In contrast, V1 and V2 demonstrate the ability to discover and classify novel categories, with varying degrees of success across languages. The OSLD methods exhibit strong performance on $\mathcal{T}_1$, with V1 achieving $84.5\%$ accuracy on Arabic. For French, both V1 and V2 models achieve good performance on $\mathcal{T}_1$. In general, the performance levels attained on $\mathcal{T}_1$ are not maintained on $\mathcal{T}_2$ and $\mathcal{T}_3$. This confirms the difficulty of learning an increasing number of new categories over time.

\noindent
\textbf{Known vs.~unknown classes.}
Comparing the results in Tables \ref{tab:results_known} and \ref{tab:results_unknown}, we observe that all models exhibit considerable performance degradation when going from known classes to unknown classes. This observation highlights the difficulty of discovering and learning new classes without supervision. Overall, the empirical results confirm that MOSLD-Bench is a challenging benchmark, and the OSLD task remains open for future work.

\vspace{-0.1cm}
\subsection{Small vs.~Large Language Models}
\label{sec_small_vs_large}
\vspace{-0.1cm}

\begin{table}[t!]
\centering
\setlength\tabcolsep{0.35em}
\resizebox{\linewidth}{!}{%
\begin{tabular}{lcccc}
\toprule
\multirow{2}{*}{\textbf{Language}} & \multicolumn{4}{c}{\textbf{Test 1}} \\
 & \textbf{Baseline} & \textbf{V1} & \textbf{V2} & \textbf{GPT-4o} \\
\midrule
French   & 0.340 / 0.178  &  \textbf{0.889} / \textbf{0.874}  & 0.862 / 0.845 & 0.713 / 0.630 \\
Turkish  & \textbf{0.518} / \textbf{0.434}  &  0.507 /  0.376  & 0.512 / 0.379  & 0.454 / 0.362 \\
\bottomrule
\end{tabular}
}
\vspace{-0.2cm}
\caption{\textbf{Overall} accuracy and F1 scores for BERT-based methods (baseline, V1 and V2) versus GPT-4o. Results are reported for the $\mathcal{T}_1$ test set. The best score for each language is highlighted in bold.}
\label{tab:gpt4o_t1}
\end{table}

We further compare the BERT-based models with a large language model \cite{Hurst-Arxiv-2024} evaluated in the OSLD setting. More precisely, GPT-4o is provided with the test data, and asked to automatically discover new classes and label test instances. This is comparatively harder than a typical zero-shot setting, where the model is given all class names. We provide the set of prompts used for GPT-4o in Table \ref{tab:prompts_gpt4o}.

The comparison between BERT-based and GPT-4o language models is presented in Table \ref{tab:gpt4o_t1}. Due to the high pricing of LLM inference, this analysis is restricted to the first evaluation stage ($\mathcal{T}_1$), and only two languages (French and Turkish).

While GPT-4o demonstrates strong OSLD capabilities, we observe a tendency towards leveraging data leakage. For Turkish, the model immediately proposes nearly the full set of original labels, suggesting reliance on prior knowledge acquired during pre-training, rather than actual discovery from the test distribution. A similar behavior is observed for French, where the model tends to predict a large number of distinct categories, indicating a bias towards over-generation that may be sourced from knowledge acquired during large-scale pre-training. Quantitatively, the proposed OSLD frameworks (V1 and V2) outperform the large language model, while operating under substantially lower computational cost. These findings highlight a trade-off between the expressive power of LLMs and the methodological alignment of smaller task-specific models.

\end{document}